\documentclass[conference]{IEEEtran}

\ifCLASSINFOpdf
   \usepackage[pdftex]{graphicx}
          \else
                  \fi

\usepackage{amsmath}

\usepackage{url}

\begin{document}
\title{Astronomical Image Reconstruction with Convolutional Neural Networks}

\author{\IEEEauthorblockN{R\'emi Flamary}
\IEEEauthorblockA{Universit\'e C\^ote d'Azur\\ Lagrange, OCA, CNRS\\ Nice, France\\
Email: remi.flamary@unice.fr
}
}

\maketitle

\begin{abstract}
State of the art methods in astronomical image reconstruction rely on
the resolution
of a regularized or constrained optimization problem. Solving this
problem can be
computationally intensive and usually leads to a
quadratic or at least superlinear complexity \emph{w.r.t.} the number
of pixels in the image. 
We investigate in this work the use of convolutional neural networks
for image reconstruction in astronomy. With neural networks, the
computationally intensive tasks is the training step, but the
prediction step has a fixed complexity per pixel, \emph{i.e.} a linear
complexity. 
Numerical experiments
show that our approach is
both computationally efficient and competitive with other state of the
art methods in addition to being interpretable.
\end{abstract}

\IEEEpeerreviewmaketitle

\section{Introduction}
\label{sec:intro}

Astronomical image observation is plagued by the fact the the observed
image is the result of a convolution between the observed object and
what the astronomers call a Point Spread Function (PSF)
\cite{starck2002deconvolution}
\cite{puetter2005digital}. In addition to the convolution the
image is also polluted by noise that is due to the
low energy of the observed objects (photon noise) or to the
sensor. The PSF is usually known \emph{a priori}, thanks to a physical
model for the telescope of estimation from known objects. 
State of the art approaches in astronomical image
reconstruction aim at solving an
optimization problem that encodes both a data fitting (with observation
and PSF) and a regularization
term that promote wanted properties in the images
\cite{starck2002deconvolution,wiaux2009compressed,theys2016reconstructing}.
Still, solving a large optimization problem for each new image can be
costly and might not be practical in the future. Indeed in the coming
years several new generations of instruments such as the Square kilometer Array
\cite{dewdney2009square} will provide very large images (both in
spatial and spectral dimensions) that will need
to be processed efficiently.

The most successful image reconstruction approaches rely on convex optimization
\cite{wiaux2009compressed,theys2016reconstructing,deguignet2016distributed}
and are all based on gradient \cite{nocedal2006numerical} or proximal splitting
gradient descent \cite{combettes2011proximal}. Interestingly those
methods have typically a linear convergence, meaning that the number of
iterations necessary to reach a given precision is proportional to the
dimension $n$ of the problem \cite{beck2009fast}, where $n$ is the
number of pixels. Since
each iteration is at best of complexity $n$ (the whole image is updated),
the overall complexity of the optimization is $O(n^2)$. Acceleration
techniques such as the one proposed by Nesterov
\cite{nesterov2005smooth}\cite{beck2009fast} manage to reduce this
complexity to $O(n^{1+1/2})$ which is still superlinear \emph{w.r.t.} the
dimension of the image. This is the main motivation for the use of
neural networks since they lead to a
linear complexity $O(n)$ that is much more tractable for large scale images.

Convolutional
neural networks (CNN) have been widely
used in machine learning and signal processing due to their impressive
performances
\cite{lecun1998gradient,krizhevsky2012imagenet,lecun2015deep}. They
are of particular interest in our case since the complexity of the
prediction of a given pixel is fixed \emph{a priori} by the
architecture of the network, which implies a linear complexity for the
whole image.
They also have been
investigated early for image reconstruction \cite{zhou1988image} and
recent advances in deep learning have shown good reconstruction performances on
natural image \cite{xu2014deep}. 

The purpose of this work is to investigate the feasibility and
performances of CNN in astronomical image
reconstruction. In the following we
first design such a neural network and discuss its learning and
implementation. Then we provide numerical experiments on real
astronomical images with a comparison to other state of the art
approaches followed by an interpretation of the learned model.

\section{Convolutional neural network for image reconstruction}
\label{sec:cnn_irec}

\begin{figure*}[t]
  \centering
  \includegraphics[width=.95\linewidth]{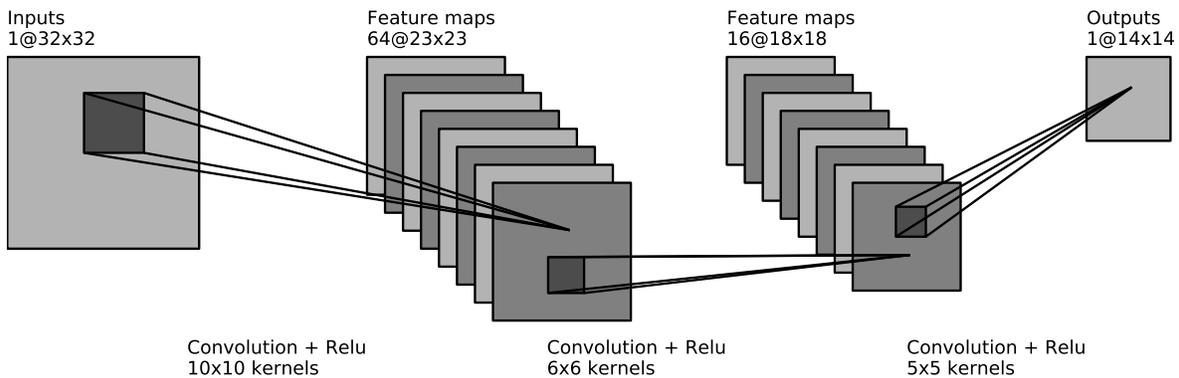}  \caption{Architecture of the proposed convolutional neural
    network. On the upper part are reported the size of the input,
    intermediary images and outputs. On the lower part the parameters
    and size of the filters for each layer.}
  \label{fig:archi}
\end{figure*}

\subsection{Network architecture}

Neural network models rely on processing the input through several layers
each consisting  of a linear
operator followed by a nonlinear transformation with an activation
function \cite{lecun2015deep}. 
Multi-layers and more
recently deep neural nets allow for a more complex nonlinear model at
the cost of a more difficult model estimation \cite{lecun2015deep}. We choose in this work to use a 3-layer
convolutional neural networks with  Rectified Linear Unit activation (ReLU)
 as illustrated in Figure
\ref{fig:archi}. We discuss in the remaining of this section the
reasons for those choices.

As their name suggests, convolutional layers perform their linear
operator as a convolution. 
They  have
been shown to work well on image reconstruction problems
in \cite{xu2014deep} which is why we use them in our design.
An interesting property is that the number of parameters of a layer
depends only on the size of the convolution (the filter) and not the
size of the input image. Also a convolution operator is common and can
benefit from hardware acceleration if necessary (Cuda GPU, DSP).
 We used $64$, $10\times 10$ filters in the first layer in order to
allow for more complex and varied filtering of the raw data to feed
the higher order representations of the following layers.
Note that at each layer the
dimensionality of the image decreases because we want the convolution
to be exact which implies a smaller output. In our model as shown in
Figure \ref{fig:archi}, we have a $14\times14$ output for a
$32\times32$ input image. From the multiple convolutions, we can see
that the predicted
value of a unique pixel is obtained from a $18\times 18$ window in the
input image. Finally this architecture has been designed for a
relatively small PSF, and the size of the filters should obviously be
adapted for large PSF.

Rectified Linear Units (ReLU) activation function has the form
$f(x)=\max(0,x)$ at each layer \cite{nair2010rectified}. ReLU is
known for its ability to train efficiently deep networks
without the need for pre-training \cite{glorot2011deep}. One strength of
ReLU is that it will naturally promote sparsity since all negative
values will be set to zero \cite{glorot2011deep}. 
Finally the maximum forces the output of each
layer to be positive, which is an elegant way to enforce the positive physical
prior in astronomical imaging.

\subsection{Model estimation}
\label{sec:model_estim}

The model described above has a number of parameters that have to be
estimated. They consist of $64$, $10\times 10$ filters for the
first layer, $64*16$, $6\times 6$ for layer two and finally $16$,
$5\times 5$ filters for the last layer that merges all the last feature
maps.  
It is interesting to note that when compared to other reconstruction
approaches, the optimization problem applies to the model
training that only
needs to be done once.
We use Stochastic Gradient Descent (SGD) with
minibatch to minimize the square Euclidean reconstruction loss. In
practice it consists in updating the parameters using at
each iteration the gradient computed only on a subset (the minibatch)
of the training samples.
This is a common approach that allows for very large
training datasets and limits the effect of the highly non convex
optimization problem \cite{krizhevsky2012imagenet}. Interestingly,
minibatches are handled in two different ways in our application. Indeed
we can see that for a given input in the model the gradient is
computed for all the $14\times 14$ output pixels, which can be seen as
a local minibatch (using neighboring pixels). One can also use a
minibatch that computes the gradient for several inputs that can come
from different places in one image and even from different
images. This last approach is necessary in practice
since it limits overfitting and helps decrease the variance of the
gradient estimate \cite{lecun2015deep}.

Finally we discuss the training dataset. Due to
the complexity of
the model and the large number of parameters, a large dataset has to
be available. We propose in this paper to use a similar approach as
the one in \cite{xu2014deep}. Since we have a known model for the PSF,
we can use clean astronomical images and generate convolved and noisy
images used to generate the dataset.
The input for a given sample is extracted from the convolved
and noisy image while the output is extracted from the clean
image. In the numerical
experiments we extract a finite number of
samples, whose positions are randomly drawn from several images. A
large dataset drawn from several images will allow
the deep neural network to learn the statistical and
spatial properties of the images and to
generalize it to unseen data (new image in our case).

\subsection{Model implementation and parameters}
\label{sec:implementation}

The model implementation has been done using the Python toolboxes
Keras+Theano that allow for a fast and scalable prototyping. They
implement the use of GPU for learning the neural network (with SGD) and
provide efficient compiled functions for predicting with the model. 
Learning a neural network requires to choose a number of parameters
that we report here for research reproducibility.
 We set the learning rate (step of the gradient descent) to
$0.01$ with a momentum of $0.9$ and we use a Nesterov-type
acceleration \cite{nesterov2005smooth}. The size of the minibatch
discussed above is set to $50$ and we limit the number of epochs
(number of times the whole dataset is scanned) to $30$. In addition we
stop the learning if the generalization error on a validation dataset
increases between two epochs.
The training dataset contains $100, 000$ samples and the validation
dataset, also obtained from the training images, contains $50, 000$
samples. Using a NVIDIA Titan X GPU, training on one epoch takes
approximately 60 seconds.

\section{Numerical experiments}

In this section we first discuss the experimental setup and then report a
numerical and visual comparison. Finally we illustrate the
interpretability of our model. Note that all the code and estimated
models will be available to the community on
GitHub.com\footnote{GitHub repository: \url{https://github.com/rflamary/AstroImageReconsCNN}}.

\subsection{Experimental setting}
\label{sec:expe_setting}

\begin{figure}[t]

  \centering
  \includegraphics[width=1\linewidth]{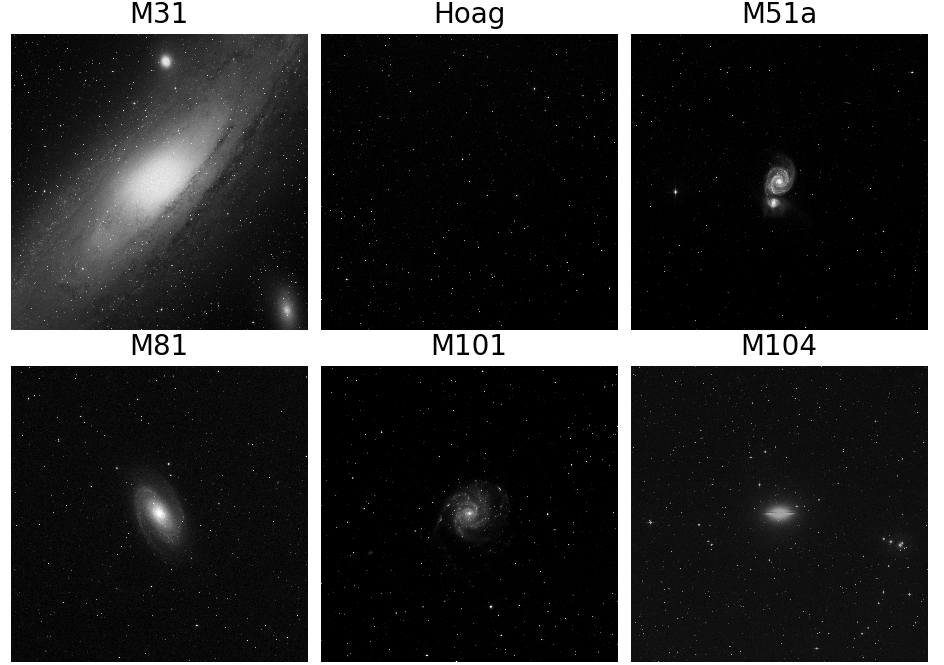}
    \caption{Images used for reconstruction comparison.}
  \label{fig:images}
\end{figure}

We use in the experiment 6 astronomical images
of well known objects
obtained from the STScI Digitized Sky Survey, HST Phase 2
dataset\footnote{Dataset website: \url{http://archive.stsci.edu/cgi-bin/dss_form}}. The
full images, illustrated in Figure \ref{fig:images}, are $3500\times
3500$ and represent a
$60$ arc minute aperture. Those images were too large for an
extensive parameter validation of the methods so we used  $1024\times
1024$ images centered on the objects for the reconstruction
comparison. The full images were used only for training the neural networks.
All images have been normalized to have a
maximum value of $1.0$. They are then convolved by a PSF with circular
symmetry corresponding
to a circular aperture telescope also known as an Airy pattern of the
form $PSF(r)=I_0 \left({J_1(r)}/{r}\right)^2$
where $r$ is a radius scaled to have a width at half maximum of $8$
pixels in a $64\times 64$ (see Figure \ref{fig:visu_com}
first row). Finally a Gaussian noise of standard
deviation $\sigma=0.01$ is added to the convolved image.

We compare our approach to a classical
Laplacian-regularized Wiener filter (Wiener)
\cite[Sect. 3]{starck2002deconvolution} and 
the iterative Richardson-Lucy
algorithm (RL) \cite{richardson1972bayesian,lucy1974iterative} that is
commonly used in astronomy. We also compare to a more modern
optimization based approach that aim at reconstructing the image using
a Total Variation regularization \cite{condat2014generic} (Prox. TV). This last
algorithm has been used with success in radio-astronomy with
additional regularization terms \cite{deguignet2016distributed}.
This last approach is a proximal splitting method with a
computational complexity equal to
\cite{wiaux2009compressed,deguignet2016distributed}. In order to limit
the computational time of the image reconstruction, Prox. TV was
limited to $100$ iterations.
Finally, in order to
evaluate the interest of using non-linearities in the layers of our
neural network, we also investigated a simple 1-layer convolutional
neural net (1-CNN) with a unique $18\times 18$ linear filter (same window as
the proposed 3-CNN model) and a linear
activation function. Interestingly this last method can also be seen
as a Wiener filtering whose components are estimated from the data.

In order to have a fair comparison we selected for all
the state of the art methods the
parameters  that maximize the reconstruction
performance on the target image. In order to evaluate the
generalization capabilities of the neural networks, the networks are
trained for each target image using samples from the $5$ other
images. In this configuration the prediction performance of the
network is evaluated only on images that were not used for training.

\subsection{Reconstruction comparisons}
\label{sec:rec_comp}

\begin{table}[t]
  \centering\small
\resizebox{.99\columnwidth}{!}{
  \begin{tabular}{|l|c|c|c|c|c|}\hline

      Image      &Wiener    &RL        &Prox. TV     &1-CNN      &3-CNN      \\\hline\hline

M31         &     35.42&     35.11&     35.17&     35.82&     \textbf{36.03} \\
Hoag        &     37.29&     37.99&     36.66&     38.58&     \textbf{39.99} \\
M51a        &     38.07&     38.26&     37.68&     38.96&     \textbf{39.99} \\
M81         &     35.91&     35.90&     35.38&     36.79&     \textbf{37.26} \\
M101        &     36.66&     37.79&     35.87&     38.31&     \textbf{39.63} \\
M104        &     36.01&     35.89&     35.34&     37.25&     \textbf{38.10} \\\hline\hline

Avg PSNR   &     36.47&     36.65&     35.93&     37.48&     \textbf{38.23}\\\hline
Avg time (s)   &      \textbf{0.24}&      1.15&    203.44&      0.52&      1.30\\\hline
  \end{tabular}}\vspace{1mm}
  \caption{Peak Signal to noise ratio (dB) for all image reconstruction
    methods on all 6 astronomical images. We also report the average
    computational time in sec. of each methods}
  \label{tab:perfs}
\end{table}

\begin{figure*}[t]
  \centering
  \includegraphics[width=\linewidth]{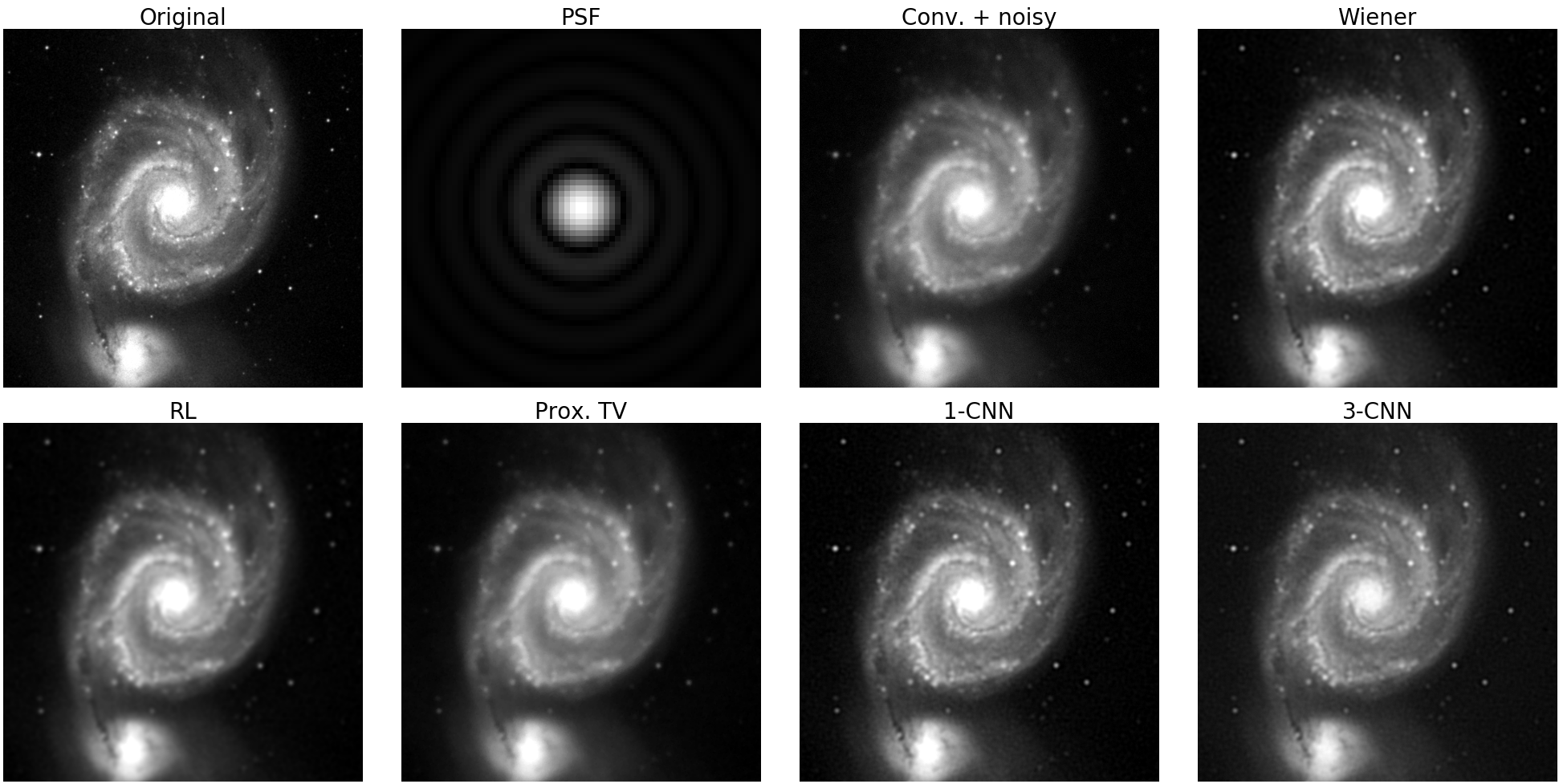}  \caption{Visual comparison of the reconstructed images for M51a. Note that
    the PSF image is zoomed and we show the square root of its
    magnitude for better illustration.}
  \label{fig:visu_com}
\end{figure*}

\begin{figure}[t]
  \centering
  \includegraphics[width=1\linewidth]{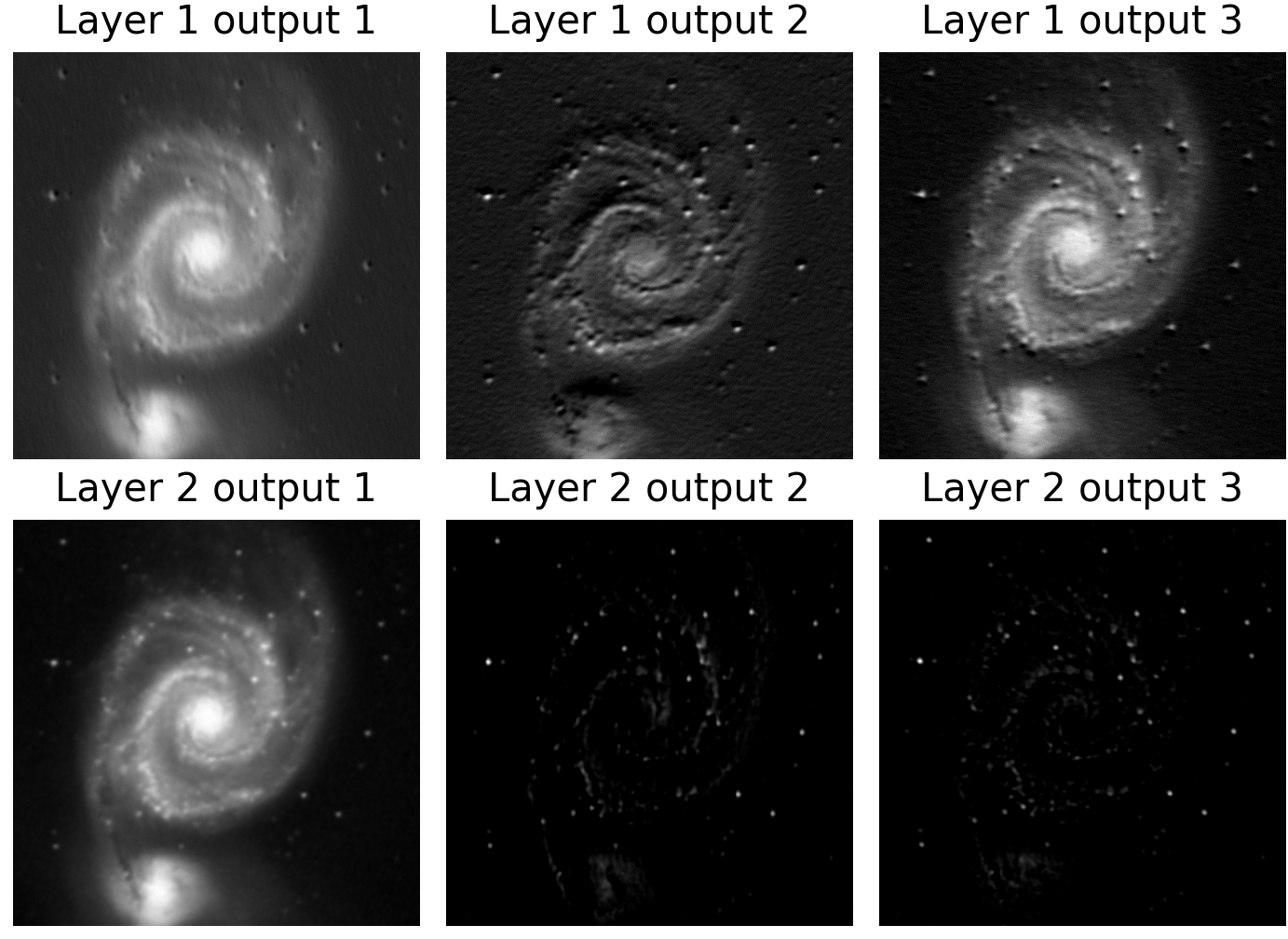}
   \caption{Selection of intermediate feature maps in the first (top
    row) and second
    (bottom row) layer of
    the neural network.}
  \label{fig:images_features}
\end{figure}

Numerical performances in Peak Signal to Noise Ratio (PSNR) are
reported  for all methods and images in Table
\ref{tab:perfs}. Note that our proposed 3-CNN neural network
consistently outperforms
other approaches while keeping a reasonable computational time
 ($\approx 1$s for a $1024\times 1024$ image on the machine used for
 the experiments). 1-CNN works
surprisingly well probably because the large dataset allows for a
robust estimation of the linear parameters. In terms of computational
complexity, Wiener is
obviously the fastest method followed by Richardson-Lucy that stops after
few iterations in the presence of noise. Our approach is in practice very
efficient compared to the optimization based approach Prox.
TV. In this scenario Prox TV leads to limited performances suggesting
the use of alternative regularizers such as Wavelets
\cite{wiaux2009compressed,deguignet2016distributed} that we will
investigate in the future. Still note that
even if other prior could lead to better performances, the
computational complexity will be the same. One of the major interest
of our method here is that its complexity is linearly proportional to
the number of pixel so that we can predict exactly the time it will
take on a given image ($1.3*4=5.2s$ on a $2048\times 2048$
image). Also since the network use a finite input window, it can be
performed in parallel on different part of the image which allow a
very large scaling on modern HPC or cloud.

Finally we also show a visual comparison of the different image
reconstructions on a small part of the M51a image in Figure
\ref{fig:visu_com}. Again we see that the 3-layer convolutional neural
network leads to a more detailed reconstruction.

\subsection{Model interpretation}
\label{sec:model_intrp}

One of the strengths of CNN is that the
feature map at each layer is an image that can be visualized. We
report in Figure \ref{fig:images_features} a selection of features
maps from the output of the first and second layers. We can see that
the first layer contains some low pass or directional filtering
whereas layer two contains more semantic information. In this case we
can see in layer 2 an image that represents
a smooth component describing the galaxy and a more high frequency
component representing point sources of a star field. The source
separation was learned by the network using only data which is very
interesting because it is a well known prior in astronomical images
that have been enforced in \cite{giovannelli2005positive}.

\section{Conclusion}

This work is a first investigation of the use of CNN for image
reconstruction in astronomy. We proposed a simple but efficient model
and numerical experiments have shown very encouraging performances
both in terms of reconstruction and computational speed. Future works
will investigate more complex PSF such as the one of
radio-interferometric telescopes
\cite{dewdney2009square,wiaux2009compressed}
and extensions to hyperspectral imaging where 3D image reconstruction
will require efficient reconstruction
\cite{deguignet2016distributed,ammanouil2017multi}. Finally the linear convolution model with noise
is limited and we will investigate datasets obtained using more
realistic simulators such as MeqTrees \cite{noordam2010meqtrees} for
radio-interferometry or CAOS \cite{carbillet2005modelling} for optical
observations.

\section*{Acknowledgment}

This work has been partly financed by ANR Magellan (ANR-14-CE23-0004-01).

\bibliographystyle{IEEEbib}

\end{document}